# Pipelined Biomedical Event Extraction Rivaling Joint Learning

Pengchao Wu, Xuefeng Li, Jinghang Gu, Longhua Qian and Guodong Zhou

**Abstract**—Biomedical event extraction is an information extraction task to obtain events from biomedical text, whose targets include the type, the trigger, and the respective arguments involved in an event. Traditional biomedical event extraction usually adopts a pipelined approach, which contains trigger identification, argument role recognition, and finally event construction either using specific rules or by machine learning. In this paper, we propose an n-ary relation extraction method based on the BERT pre-training model to construct Binding events, in order to capture the semantic information about an event's context and its participants. The experimental results show that our method achieves promising results on the GE11 and GE13 corpora of the BioNLP shared task with F1 scores of 63.14% and 59.40%, respectively. It demonstrates that by significantly improving the performance of Binding events, the overall performance of the pipelined event extraction approach or even exceeds those of current joint learning methods.

**Index Terms**—Biomedical event extraction, BERT, N-ary relation extraction, Pipeline

✦ —————————

## 1 INTRODUCTION

Information extraction in the biomedical field focuses on how to automatically extract useful information for researchers from a huge amount of biomedical texts and presents the biomedical knowledge in a structured form [1]. This structured information includes biomedical entities, relations between entities and events, etc., which have important applications and research significance in building pathways [2] and enriching databases [3]. Biomedical events are changes in the state of one or more entities, such as *Gene Expression*, *Transcription*, *Phosphorylation*, *Regulation*, etc. Events have their specific types, triggers and arguments, where the triggers are used to identify the occurrence of an event and the arguments are the participants with specific roles.

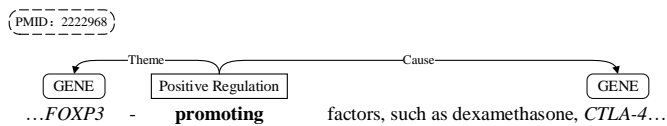

Fig. 1. An event example.

Fig. 1 shows an example of an event from the GE11 corpus. The text contains a *Positive Regulation* event, which is triggered by the word "**promoting**" and involves two arguments: one is the *Theme* argument "*FOXP3*", and the other is *Cause* argument "*CTLA-4*".

Compared with event extraction in other domains, biomedical event extraction suffers from challenges such as ambiguity in triggers, flexibility in the number of event arguments and the presence of nested events, which indicates that biomedical event extraction is a challenging research direction. The mainstream frameworks for biomedical event extraction can be basically divided into two categories: pipelined and joint learning.

The pipelined approach decomposes the event extraction task into three successive sub-tasks, first identifying triggers from the text, then recognizing the arguments of these triggers and their corresponding roles, and finally combining the identified trigger and its arguments to obtain the final event. The pipelined approach can be further subdivided into three types: 1) **Rule-based methods**, which generally use dictionary matching to identify triggers, and then uses dependency rules [4], semantic rules [5] or structured templates [6], [7] to identify the arguments, and finally construct the events according to the constraints on the number and roles of different event types. 2) **Machine learning-based methods**, which use lexical, syntactic and semantic-based features to automatically identify triggers and determine the role of the arguments relative to the triggers. Most of machine learning methods use SVM classifiers [8], [9], [10], [11], [12]. Event construction can be performed using rules [8], [12] or by classifiers to determine the validity of a potential event [9], [10]. 3) **Deep learning-based methods**, which use neural network models to encode and classify the input text to identify triggers and determine the role of relations between triggers and arguments. In addition to text, the input can also include syntactic and semantic information, and the network models can be CNN [13] [14], [15], or Tree-LSTM [16] based on dependency trees [17].

---

- *P.C. Wu is with the School of Computer Science and Technology, Soochow University, Suzhou, Jiangsu 215006, China. E-mail: 20204227037@stu.suda.edu.cn.*
- *X.F. Li is with the School of Computer Science and Technology, Soochow University, Suzhou, Jiangsu 215006, China. E-mail: 20215227070@stu.suda.edu.cn.*
- *J.H. Gu is with the Department of Chinese and Bilingual Studies, The Hong Kong Polytechnic University, Hong Kong 999077, China. E-mail: gujinghangnlp@gmail.com.*
- *L.H. Qian is with the School of Computer Science and Technology, Soochow University, Suzhou, Jiangsu 215006, China. E-mail: qianlonghua@suda.edu.cn.*
- *G.D. Zhou is with the School of Computer Science and Technology, Soochow University, Suzhou, Jiangsu 215006, China. E-mail: gdzhou@suda.edu.cn.*

One disadvantage of the pipelined approach is that errors occurring in the upstream tasks will be propagated to the downstream tasks, leading to cascading errors. The joint learning approach, on the contrary, combines multiple related tasks to learn jointly, aiming to overcome the shortcomings of the pipelined approach. **Parameter sharing** is one common way of joint learning, in which multiple tasks share part of the network and its parameters. DeepEventMine [18] uses SciBERT [19] model to encode the text and then adopts a span-based method to identify triggers, the role of arguments with the triggers and finally determine whether the combinations of a trigger and its arguments constitute valid events. The objective functions of these three tasks are combined in order to jointly train the model. Huang et al. [20] also use SciBERT to encode a sentence, then incorporate knowledge of concepts and relations from UMLS to form a graph structure, which is encoded by a GNN model. The graph's output is used to perform trigger identification and argument extraction, however, events are constructed using a rule-based way. Zhao et al. [21] proposed an end-to-end document-level event extraction framework that models the interaction between local and global contexts through stacked HANN layers, then feeds the final representations into a joint extraction layer to identify triggers and the role of arguments using a sequence labeling-based method. Wang et al. [22] use the BioBERT [23] model to encode the text, obtain the contextual representations using BiLSTM, model the dependency information of the sentences using GCN, fuse the two types of information into the joint extraction layer for the trigger classifier and event classifier of biomedical events. To address the sparsity of annotated data, Zhao et al. [24] also use BioBERT to encode the text, but perform data augmentation by a self-supervised learning-based method, and use reinforcement learning for sequence labeling-based trigger identification and argument extraction. Ramponi et al. [25] encode the sentence with BERT [26] and use a **multi-label** approach to identify whether a word is an event trigger, or if it is an entity, what role it has with other triggers. **Machine reading comprehension** provides a new way for joint learning. Wang et al. [27] convert biomedical event extraction into an entity-driven multi-turn question answering task to sequentially identify the trigger corresponding to an entity, the other arguments of the trigger. The method allows for events with multiple arguments and nested events.

In general, joint learning models are complex and relatively difficult to train, on the other hand, with the widespread use of powerful pre-trained language models, the pipelined approach also has the potential to achieve better performance for NLP tasks. In this paper, the BioBERT model is used as the base encoder to sequentially perform trigger identification, argument role recognition and final event construction in a pipelined manner. To address the errors in event construction, especially the false positives of *Binding* events whose types are "*Binding*", we propose a construction method for *Binding* events based on n-ary relation extraction, which not only significantly improves the extraction performance of *Binding* events, but also promotes the performance of regulation events (*Regu*, *PoRe*, *NeRe*) that depend on *Binding* events, and to some extent alleviates the cascading errors brought by the pipelined approach, making our pipelined approach rivaling joint learning models.

## 2 PIPELINED BIOMEDICAL EVENT EXTRACTION

We adopt a pipelined approach for biomedical event extraction, which is divided into three successive sub-tasks, including trigger identification, argument role recognition and event construction, and the overall flow is shown in Fig. 2. The trigger identification task identifies the location and type of the event trigger from the text, while the argument role recognition task recognizes the arguments and their roles associated with the trigger, and finally the event construction task uses either rule-based or machine learning methods to combine the trigger and its arguments into an event.

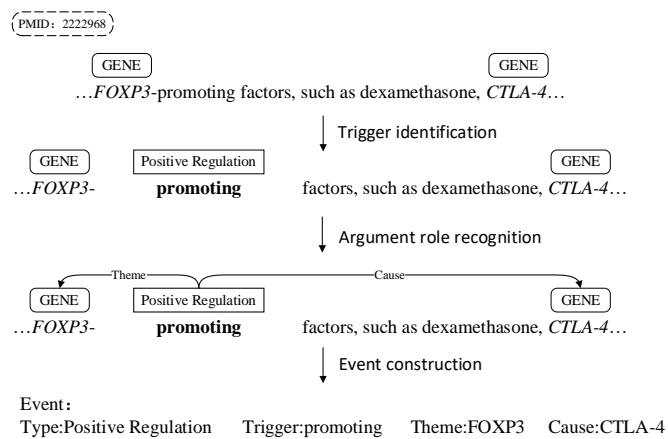

Fig. 2. Flow chart of the pipelined biomedical event extraction.

In Fig. 2, given the input text "...*FOXP3*-promoting factors, such as dexamethasone, *CTLA-4*..." and the entities' locations and types, the *Positive Regulation* trigger "**promoting**" is identified by the sub-task of trigger identification; In the sub-task of argument role recognition, the identified trigger is paired with the given two entities for argument role recognition, and thus a *Theme* argument "*FOXP3*" and a *Cause* argument "*CTLA-4*" are obtained; finally during event construction, an event is obtained by combination of the trigger and two arguments.

### 2.1 Trigger Identification

Like named entity recognition (NER), the trigger identification task can be regarded as a sequence labeling task, and we adopt a BERT-based model to perform the identification of trigger positions and types using the label schema of BIO as illustrated in Fig. 3. To highlight the role of gene entities, their occurrences in the sentence are masked and replaced with entity type "gene".

### 2.2 Argument Role Recognition

Argument role recognition is to find the entities that have an argument-role relation with the trigger and can be regarded as a relation classification task with the classes of *Theme*, *Cause* and *None*. As with relation extraction, a BERT-based classification method is used. The trigger is the subject of the relation, and the argument is the object. It should be noted that since the corpus contains nested events, the argument of one event may be another sub-event, therefore the object may be an entity or another trigger indicating another event. To be able to distinguish the trigger and the argument in the sentence, special symbols are used to mark the trigger and the argument. The trigger is marked around with "#" while the argument is marked around with "@". The entity is still masked with the entity type "gene", the trigger, however, remains intact as shown in Fig. 4.

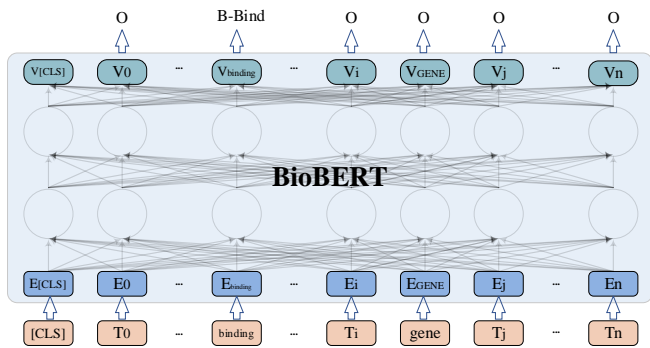

Fig. 3. Trigger identification model.

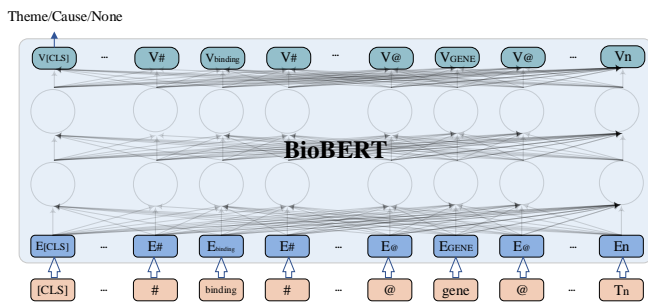

Fig. 4. Argument role recognition model.

**2.3 Rule-based event construction**

After obtaining the triggers and arguments in the sentences, a rule-based approach can be used to construct the events. This subsection first introduces the argument composition for different types of events, then introduces the corresponding construction rules based on the argument composition, and finally summarizes the problems in the rule-based event construction.

1) *Argument composition for biomedical events*

Table 1 shows the argument composition for different event types in the GE11[28] and GE13[29], [30] corpora, where the event types in boldface are specific to the GE13 corpus. As seen from the Table 1:

1. *GeEx*, *Tran*, *PrCa*, *Phos*, *Loca* have a simple argument composition, with only one *Theme* entity argument, called **simple events.**
2. *Bind* events only have *Theme* entity arguments, but the number can be one or more, so they are called **multiple events**.
3. *Regu*, *PoRe*, *NeRe*, *PrMo*, *Ubiq*, *Acet*, *Deac* have the most *complex composition of the argu*ment, which must contain a *Theme* argument, and may contain 0~1 *Cause* argument, and the argument of both roles of the regulation events may be entities or events, so these events are called **nested events**.

2) *Construction rules*

According to the argument composition of events, most of them can be successfully constructed with specific rules as follows:

1. **Isolated triggers:** If the trigger does not have any *Theme* arguments, no event is generated.
2. **Simple events:** the trigger and each of its *Theme* arguments form a separate event.
3. **Multiple events:** For *Binding* events, if the trigger has only one *Theme* argument, it constitutes an event; when the number of *Theme* arguments is greater than or equal to two, the arguments are paired with each other and then combined with the trigger to form respective events.
4. **Nested events:** If the trigger has only *Theme* arguments, the trigger and each *Theme* argument is assembled into an event; if it contains both *Theme* and *Cause* arguments, each of *Theme* arguments is paired with each of *Cause* arguments to form an event. Note that since an argument may be an event indicated by another trigger, we use a recursive bottom-up method to construct the low layer events first, and then generate the nested events upwards.

3) *Problems in rule-based event construction*

While the rule-based event construction is simple and efficient, with high accuracy for simple events and considerable performance for nested events, for multiple events like *Binding*, its performance is unsatisfactory. The reason is that there is no reliable combination rule to determine whether a *Binding* trigger and its multiple *Theme* arguments constitute an event, rather it depends on the semantics expressed both in the sentence and the potential event. A simple pairwise method of *Theme* arguments will produce false positives along with false negatives. In the sentence shown in Fig. 5, the trigger "**bind**" has three *Theme* arguments (*Sp1*, *Sp3* and *A3G*), and these three arguments are not paired to generate three events, rather *Sp1* and *A3G*, *Sp3* and *A3G* are paired to generate two *Binding* events respectively.

TABLE 1
COMPOSITION OF ARGUMENTS FOR DIFFERENT TYPES OF EVENTS

| Event Types | Abbr. | Composition of arguments |
|---|---|---|
| Gene Expression | GeEx | Theme(1), Entity |
| Transcription | Tran | Theme(1), Entity |
| Protein Catabolism | PrCa | Theme(1), Entity |
| Phosphorylation[a] | Phos | Theme(1), Entity |
| Localization | Loca | Theme(1), Entity |
| Binding | Bind | Theme(1 and more), all entities |
| Regulation | Regu | Theme(1), Cause(0~1), Entity or event |
| Positive Regulation | PoRe | Theme(1), Cause(0~1), Entity or event |
| Negative Regulation | NeRe | Theme(1), Cause(0~1), Entity or event |
| **Protein Modification** | PrMo | Theme(1), Entity, Cause(0~1), Entity or event |
| **Ubiquitination** | Ubiq | Theme(1), Entity, Cause(0~1), Entity or event |
| **Acetylation** | Acet | Theme(1), Entity, Cause(0~1), Entity or event |
| **Deacetylation** | Deac | Theme(1), Entity, Cause(0~1), Entity or event |

[a]*Phosphorylation events differ in the composition of the arguments in GE11 and GE13, but the number of events in GE13's form is smaller. To unify the processing, we use the composition of arguments in the GE11 corpus as a standard.*

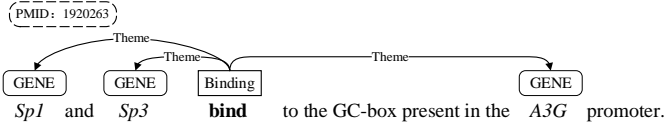

Fig. 5. Examples of errors in rule-based *Binding* event construction.

## 3 AUTOMATIC EVENT CONSTRUCTION BASED ON N-ARY RELATION EXTRACTION

On the one hand, the rule-based event construction makes the performance of the *Binding* events unsatisfactory, on the other hand, the arguments in the regulation events (*Regulation*, *Positive Regulation* and *Negative Regulation*) may also be sub-events of *Binding*, so the performance of the *Binding* events will indirectly affect the performance of the regulation events. Due to the fact that the construction of *Binding* events is based on evidence of sentence, we cast the construction of *Binding* events into a machine learning classification problem, and further make full use of the powerful BERT model to improve its performance. Specifically, we propose an n-ary relation extraction approach to determine the validity of a potential *Binding* event. N-ary relation extraction aims to extract relations among *n* entities in the sentence context, and here we consider the trigger and the arguments as multiple entities in relation extraction. All possible combinations of triggers and arguments are treated as event candidates and they are classified into valid and invalid ones.

### 3.1 Binding event construction model

Fig. 6 shows the framework of the model used for candidate event classification, which consists of an input layer, a BERT encoder and an output layer.

In the input layer, special symbols are used to mark trigger and arguments. For the input sequence $T = \{[CLS], tok_1, ..., tok_{i-1}, tok_i, tok_{i+1}, ..., tok_n\}$, where $tok_i$ is the i-th token in the input context and [CLS] is the special token used for n-ary relation extraction classification output. The trigger is marked with "#" symbol, and the arguments are marked with "@". Similar to the previous two sub-tasks, gene entities are still masked by "gene". It is important that the other arguments of the trigger not involved in the candidate event also have an important impact and therefore need to be marked with special symbol "$". As shown in Fig. 6, among the three arguments of the trigger "**bind**", gene 1 and gene 3 are involved in the candidate event, while gene 2 is not. For a sequence of input vectors $E = \{E_{[CLS]}, E_1, ..., E_{i-1}, E_i, E_{i+1}, ..., E_n\}$, where $E_i$ is the input vector corresponding to the i-th token.

In the BERT encoder, the model obtains the output vector sequence by encoding the input vector sequence $E$ through Transformers. For a sequence of output vectors $V = \{V_{[CLS]}, V_1, ..., V_{i-1}, V_i, V_{i+1}, ..., V_n\}$, where $V_i$ is the output vector corresponding to the i-th token.

In the output layer, the probability set $p$ of candidate types is obtained by adding a fully connected layer and Softmax to the output vector $V_{[CLS]}$ of [CLS], and the class corresponding to the largest probability $\hat{y}$ in the probability set $p$ is selected as the result of the final classification. The probability set and the largest probability can be expressed as follows:

$$p = softmax(WV_{[CLS]} + b) \quad (1)$$

$$\hat{y} = argmax(p) \quad (2)$$

where W and b are the weights and bias.

The loss function $L$ for model training uses the cross-entropy loss function, as shown below.

$$L = -\sum_{i=1}^{n} \hat{y} \log \hat{y} - (1-\hat{y}) \log(1-\hat{y}) \quad (3)$$

### 3.2 Binding event instance generation

Before training or prediction, the arguments of the *Binding* events must be combinatorially combined with each other to generate candidate events. Each candidate instance contains three elements: 1) trigger; 2) participating arguments: the arguments that have the relations with the trigger and participate in the candidate event; 3) non-participating arguments: the arguments that have the relations with the trigger but do not participate in the candidate event. The steps for generating candidate instances are as follows:

1. **Prepare instance elements:** find the *Binding* event trigger and the arguments that have a *Theme* relationship with the trigger.
2. **Generate candidate instances:** enumerate all combinations between the trigger and the arguments to generate candidate instances.
3. **Determine instance labels:** in training, if the candidate instance can be found in the annotated events, it is labeled as positive, otherwise it is negative; In prediction, if the predicted result is positive, a *Binding* event is obtained.

Fig. 7 shows the process of generating n-ary relation extraction instances, where the original text contains the trigger "**bind**", its three arguments, and two annotated *Binding* events, E1 and E2. There are seven possible combinations of three arguments to generate instances, and seven candidate instances will be generated, two of which are positive instances, and the rest are negative instances.

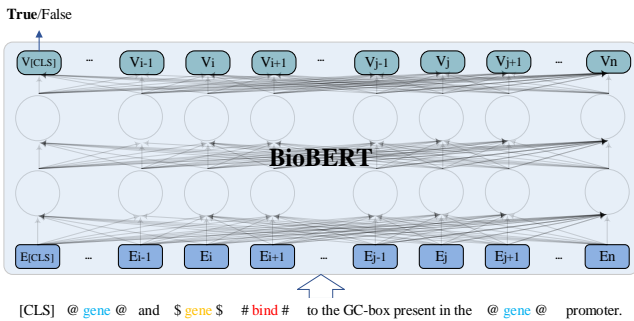

Fig. 6. *Binding* event classification model.

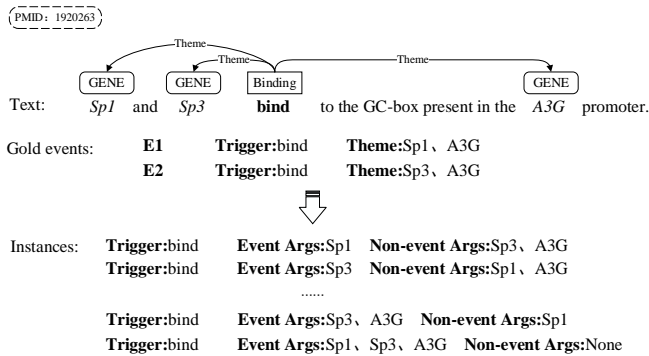

Fig. 7. Flow chart of generating n-ary relation extraction instance.

# 4 EXPERIMENTATION
## 4.1 Datasets

The commonly used corpora in biomedical event extraction are GE11 and GE13. Table 2 shows the numbers and ratios of different types of events in the two corpora, where the training and development sets are de-duplicated, while those of the test sets are derived from the official documents. As can be seen form the Table 2 that:

1. The numbers and ratios of different event types vary greatly. The four types of events (*PrMo*, *Ubiq*, *Acet* and *Deac*) unique to GE13 corpora account for less than 1% of the training/development/test sets, and some of them are even zero; while the largest numbers of *PoRe* events account for more than 27% in both corpora, and overall, the regulation events account for a large proportion.
2. The numbers of events of the same type across two corpora are roughly similar in proportions, which indicates that the contributions of different event types to the overall performance across two corpora are approximate.
3. *Binding* events account for about 10% of two corpora. Considering the largest proportion of regulation events whose arguments may be *Binding* events, it is supposed that *Binding* events have a great influence on the overall performance of event extraction.

## 4.2 Evaluation Metrics

BioNLP shared tasks provide an online evaluation tool which is used to evaluate the submitted results on the GE11/GE13 test set. It adopts the evaluation strategy of approximate span/approximate recursive, where a prediction span is considered correct if it is within one word of the annotated span, and a nested event is considered correct even if the sub-event is partially correct. However, when we evaluate the models on the development sets, all spans and events are strictly matched. The evaluation metrics of P/R/F1 are used, i.e., precision, recall and F1 score.

## 4.3 Parameter Setting

Table 3 lists the hyper-parameters used by the models in the experiments. The models are trained for 20 epochs and the best models are selected based on their performance scores evaluated on the development set.

TABLE 3
MODEL HYPER-PARAMETERS SETTING

| Hyper-parameters | Value |
|---|---|
| Model | Biobert-pubmed-v1.1 |
| Learning rate | 1e-5 |
| Batch size | 8 |
| Epoch | 5/20 |
| Max_seq_len | 256 |
| Loss function | Categorical Crossentropy |
| Optimizer | Adam |

TABLE 2
STATISTICS OF EVENTS IN GE11/GE13 CORPORA

| Type | GE11 | | | | | | GE13 | | | | | |
|---|---|---|---|---|---|---|---|---|---|---|---|---|
| | Train | | Dev | | Test | | Train | | Dev | | Test | |
| | # | % | # | % | # | % | # | % | # | % | # | % |
| GeEx | 2,265 | 22.0 | 749 | 23.1 | 1,002 | 22.3 | 729 | 26.1 | 591 | 18.5 | 619 | 18.5 |
| Tran | 667 | 6.5 | 158 | 4.9 | 174 | 3.9 | 122 | 4.4 | 98 | 3.1 | 101 | 3.0 |
| PrCa | 110 | 1.1 | 23 | 0.7 | 15 | 0.3 | 23 | 0.8 | 30 | 0.9 | 14 | 0.4 |
| Phos | 188 | 1.8 | 111 | 3.4 | 189 | 4.2 | 107 | 3.8 | 193 | 6.1 | 161 | 4.8 |
| Loca | 279 | 2.7 | 67 | 2.1 | 191 | 4.3 | 44 | 1.6 | 197 | 6.2 | 99 | 3.0 |
| Bind | 977 | 9.5 | 373 | 11.5 | 502 | 11.2 | 191 | 6.8 | 373 | 11.7 | 342 | 10.2 |
| **PrMo** | - | - | - | - | - | - | 8 | 0.3 | 1 | 0.0 | 1 | 0.0 |
| **Ubiq** | - | - | - | - | - | - | 4 | 0.1 | 1 | 0.0 | 30 | 0.9 |
| **Acet** | - | - | - | - | - | - | 0 | 0.0 | 2 | 0.1 | 0 | 0.0 |
| **Deac** | - | - | - | - | - | - | 0 | 0.0 | 4 | 0.1 | 0 | 0.0 |
| Regu | 1,112 | 10.8 | 293 | 9.0 | 388 | 8.6 | 298 | 10.6 | 284 | 9.0 | 299 | 8.9 |
| PoRe | 3,384 | 32.9 | 999 | 30.8 | 1,453 | 32.4 | 779 | 27.8 | 883 | 27.7 | 1144 | 34.2 |
| NeRe | 1,309 | 12.7 | 471 | 14.5 | 573 | 12.8 | 496 | 17.7 | 531 | 16.7 | 538 | 16.1 |
| All | 10,291 | 100 | 3,244 | 100 | 4,487 | 100 | 2,801 | 100 | 3,188 | 100 | 3,348 | 100 |

6## 4.4 Experimental Results

1) *Performance of rule-based and automatic construction in different scenarios*

Table 4 compares the overall performance of event extraction on GE11 and GE13 development sets for different scenarios with different event construction methods. The scenarios refer to whether gold or automatic triggers and arguments are used, where "Gold" indicates that the annotated triggers or arguments are used, and "Auto" indicates that the triggers or arguments are automatically identified by the BERT models. Note that when the trigger is automatic, the arguments are further recognized based on the automatic trigger. In columns, "Rule" means that the event construction is performed entirely by the rule-based approach mentioned above, and "Auto" means that the *Binding* events are extracted by the n-ary relation extraction approach while the other types are still constructed by the rule-based approach. The overall results are the average of five random runs, and the values in the parentheses besides the F1 scores are the standard variance across five runs. The higher performance scores among "Rule" and "Auto" are highlighted in boldface. Note that there is only one result when using the rule-based approach under gold triggers and arguments. As can be seen in the Table 4:

1. In all scenarios, the F1 scores of the automatic approach outperform the rule-based approach on both corpora, mainly due to the significant improvement in precision, which suggests that the *Binding* event construction based on n-ary relation extraction significantly improves the overall performance of event extraction by reducing false positives from *Binding* event construction.
2. With the gold triggers and arguments, the automatic construction approach obtains a remarkable improvement of F1 values by 2~3 points on both corpora. Obviously, with the gradual introduction of noise in the automatic triggers and arguments, the performance improvement of the automatic approach consistently decreases.

2) *Performance of rule-based and automatic construction in different scenarios*

Table 5 compares the performance of different event construction approaches on different types of events on both GE11 and GE13 corpora, where both triggers and arguments are gold and other settings are the same as in Table 5. In particular, "~" in the cells means that the results of the automatic approach are the same as those of the rule-based approach, and "-" means that there are no results for such event types.

TABLE 4
PERFORMANCE OF RULE-BASED AND AUTOMATIC CONSTRUCTION IN DIFFERENT SCENARIOS

| Corpus | Trigger | Arguments | Rule | | | Auto | | |
|---|---|---|---|---|---|---|---|---|
| | | | *P(%)* | *R(%)* | *F1(%)* | *P(%)* | *R(%)* | *F1(%)* |
| GE11 | Gold | Gold | 90.77 | 90.97 | 90.87 | **95.90** | **92.64** | **94.24**(±0.17) |
| | Gold | Auto | 75.99 | 73.39 | 74.64(±0.29) | **78.87** | **74.62** | **76.66**(±0.35) |
| | Auto | Auto | 57.72 | 54.98 | 56.23(±0.55) | **60.14** | **55.75** | **57.79**(±0.32) |
| GE13 | Gold | Gold | 90.26 | **93.32** | 91.76 | **96.18** | 92.57 | **94.33**(±0.90) |
| | Gold | Auto | 69.15 | **70.83** | 69.96(±0.70) | **72.77** | 70.16 | **71.42**(±0.79) |
| | Auto | Auto | 55.21 | **52.60** | 53.82(±1.02) | **57.62** | 52.16 | **54.71**(±0.98) |

TABLE 5
PERFORMANCE OF RULE-BASED AND AUTOMATIC CONSTRUCTION IN DIFFERENT EVENT TYPES

| Type | GE11 | | | | | | GE13 | | | | | |
|---|---|---|---|---|---|---|---|---|---|---|---|---|
| | Rule | | | Auto | | | Rule | | | Auto | | |
| | *P(%)* | *R(%)* | *F1(%)* | *P(%)* | *R(%)* | *F1(%)* | *P(%)* | *R(%)* | *F1(%)* | *P(%)* | *R(%)* | *F1(%)* |
| GeEx | 100 | 98.00 | 98.99 | ~ | ~ | ~ | 100 | 98.65 | 99.32 | ~ | ~ | ~ |
| Tran | 100 | 94.3 | 97.07 | ~ | ~ | ~ | 100 | 90.82 | 95.19 | ~ | ~ | ~ |
| PrCa | 100 | 100 | 100 | ~ | ~ | ~ | 100 | 100 | 100 | ~ | ~ | ~ |
| Phos | 100 | 94.59 | 97.22 | ~ | ~ | ~ | 93.51 | 89.64 | 91.53 | ~ | ~ | ~ |
| Loca | 100 | 92.54 | 96.12 | ~ | ~ | ~ | 100 | 97.46 | 98.71 | ~ | ~ | ~ |
| Bind | 64.85 | 70.24 | 67.44 | **92.88** | **83.91** | **88.17** | 70.64 | 85.79 | 77.48 | **97.08** | 80.16 | **87.81** |
| PrMo | - | - | - | - | - | - | 100 | 100 | 100 | ~ | ~ | ~ |
| Ubiq | - | - | - | - | - | - | 100 | 100 | 100 | ~ | ~ | ~ |
| Acet | - | - | - | - | - | - | 100 | 100 | 100 | ~ | ~ | ~ |
| Deac | - | - | - | - | - | - | 66.67 | 50.00 | 57.14 | ~ | ~ | ~ |
| Regu | 95.36 | 91.13 | 93.19 | **96.42** | **91.81** | **94.06** | 96.00 | 92.96 | 94.45 | **96.01** | **93.31** | **94.64** |
| PoRe | 86.92 | 90.49 | 88.67 | **91.28** | **91.19** | **91.24** | 83.25 | 90.60 | 86.77 | **98.64** | 90.03 | **94.14** |
| NeRe | 98.02 | 94.48 | 96.22 | **98.67** | 94.27 | **96.42** | 99.42 | 97.55 | 98.48 | **99.61** | 97.36 | **98.48** |
| Avg. | 90.77 | 90.97 | 90.87 | **95.95** | **92.79** | **94.34** | 90.26 | 93.32 | 91.76 | **98.46** | 92.50 | **95.39** |

We can see from the Table 5 that, when compared the automatic event construction with the rule one:

1. There is a significant improvement in the performance of *Binding* events, by 21 and 10 units in the F1 scores of *Binding* events in GE11 and GE13, respectively. This indicates that the n-ary relation extraction can greatly improve the construction of *Binding* events mainly due to the significant improvement in precision.
2. The performance of regulation events (*Regu*, *PoRe*, *NeRe*) is also improved to some extent, due to the fact that regulation events may contain *Binding* events as their arguments, therefore the performance improvement of *Binding* events can indirectly improve the performance of regulation events. Among them, *Binding* events occur most often as sub-event arguments in *PoRe* events, so the *PoRe* events have the largest promotion.

*3) Comparison of performance with other SOTA systems*

Table 6 compares the performance of our approach with those of others on the test sets of GE11 and GE13. As can be seen from the Table 6 that:

1. Most of the deep learning methods have higher performance than rule-based and conventional machine learning methods, suggesting that deep learning methods may capture semantic features that can better perform event extraction tasks.
2. When all events are constructed with rules, our overall performance is already competitive, and the performance on both corpora surpasses other pipelined systems in Table 6 by more than 2 units of F scores, suggesting that the powerful BioBERT model can bring an overall gain to the pipelined approach by improving the performance of respective subtasks.
3. Using the n-ary relation extraction approach to construct *Binding* events induces performance improvement of about 2 units on both corpora, indicating that our approach is also effective in improving the overall performance of event extraction on the test sets.
4. The performance of most pipelined approaches is lower than that of joint learning approaches, but with our method of event construction, the performance scores on the GE11 and GE13 corpora exceed those of best-performing joint learning models such as DeepEventMine[18] and Zhao et al.[24], suggesting that our method of event construction can also be potentially used to improve the performance of joint learning models.

## 5 DISCUSSION & ANALYSIS

### 5.1 Precision increase of Binding event extraction

As can be seen from Table 5, the precision scores of *Binding* event extraction on both corpora under the automatic approach are significantly improved compared with the rule-based one, mainly because the automatic approach can significantly reduce the number of false positives of *Binding* events in sentences with coordinating conjunctions and the number of the cascading errors for *Binding* events.

For the first example, in Fig. 8 (a), *Oct1* and *Oct2* are two conjuncts of entities, and constitute two respective *Binding* events with *BOB.1*. However, the rule-based pairwise approach constructs a false event with *Oct1* and *Oct2* as two arguments; in Fig. 8 (b), since *NF-kappaB* is a protein family, the trigger "**binding**" constitutes three respective *Binding* events with each argument (*p65*, *p50* and *FLAG-Tat*), while rule-based approach constructs three false positives with pairwise arguments.

For the second example, among all the event types of this task, the cascading errors of *Binding* events are more serious, since in the rule-based approach, each *Binding* trigger needs to be combined with arguments in different combinations, and if the trigger or the argument role is incorrectly predicted, it will lead to more errors in the subsequent subtasks of event construction. Fig. 9(a) shows the cascading errors where the triggers are incorrectly predicted, and the subsequently identified arguments are wrong, therefore multiple error events are generated during the rule-based approach. This kind of errors account for ~78% in the GE11 development set and ~60% in the GE13 development set of all cascading errors for *Binding* events, respectively. However, with our method, ~59% (GE11) and ~79% (GE13) of such erroneous events can be eliminated. Fig. 9(b) shows another example that the argument role is incorrectly identified, when the trigger prediction is correct, and similar to the former, multiple erroneous events are generated during the rule-based approach. This kind of errors accounts for ~22% (GE11) and ~40% (GE13) of all cascading errors for *Binding* events, respectively. With our method, ~55% (GE11) and ~69% (GE13) of erroneous events can be avoided.

TABLE 6
COMPARISON OF PERFORMANCE WITH OTHER SOTA SYSTEMS ON GE11 AND GE13 TEST SETS

| Method | Systems | Model | Framework | Construction | GE11 | | | GE13 | | |
|---|---|---|---|---|---|---|---|---|---|---|
| | | | | | *P(%)* | *R(%)* | *F1(%)* | *P(%)* | *R(%)* | *F1(%)* |
| Rule-based | Kilicoglu et al. [5] (2011) | - | Pipeline | Rule | 59.58 | 43.55 | 50.32 | - | - | - |
| | Bui et al. [6] (2012) | - | Pipeline | Rule | 66.63 | 44.47 | 53.34 | - | - | - |
| | BioSem [7] (2013) | - | Pipeline | Rule | - | - | - | 62.83 | 42.47 | 50.68 |
| Machine learning | Björne et al. [10] (2012) | SVM | Pipeline | Auto | - | - | 53.30 | - | - | - |
| | BioMLN [11] (2014) | SVM; MLN | Pipeline | Auto | 63.61 | 53.42 | 58.07 | 59.24 | 48.95 | 53.61 |
| | Majumder et al. [12] (2016) | SVM | Pipeline | Rule | 66.46 | 48.96 | 56.38 | - | - | - |
| Deep learning | Björne et al. [15] (2018) | CNN | Pipeline | Auto | 64.86 | 50.53 | 56.80 | 58.95 | 40.29 | 47.87 |
| | Li et al. [17] (2019) | Tree-LSTM | Pipeline | Rule | 67.01 | 52.14 | 58.65 | - | - | - |
| | DeepEventMine [18] (2020) | SciBERT | Joint learning | Auto | 71.71 | 56.20 | 63.02 | 60.98 | 49.80 | 54.83 |
| | GEANet [20] (2020) | SciBERT;GNN | Joint learning | Rule | 56.11 | 64.61 | 60.06 | - | - | - |
| | BEESL [25] (2020) | BioBERT | Joint learning | Rule | 69.72 | 53.00 | 60.22 | - | - | - |
| | QA with BERT [27] (2020) | SciBERT | Joint learning | Rule | 59.33 | 57.37 | 58.33 | - | - | - |
| | Zhao et al. [21] (2021) | HANN | Joint learning | - | 71.73 | 53.21 | 61.10 | - | - | - |
| | Zhao et al. [24] (2021) | BioBERT | Joint learning | Auto | - | - | - | 64.21 | 53.77 | 58.53 |
| | CPJE [22] (2022) | BioBERT | Joint learning | Auto | 72.62 | 53.33 | 61.50 | - | - | - |
| | Ours (Rule) | BioBERT | Pipeline | Rule | 64.93 | 57.73 | 61.12 | 59.66 | 55.38 | 57.44 |
| | Ours (Auto) | BioBERT | Pipeline | Auto | 67.04 | 59.66 | **63.14** | 63.90 | 55.50 | **59.40** |

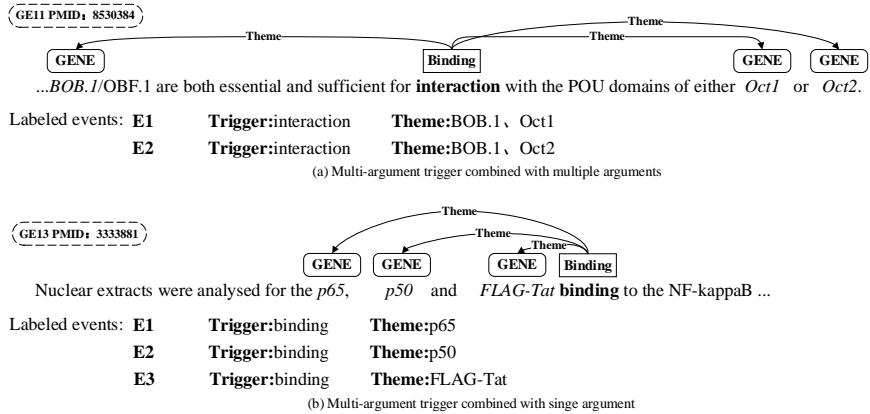

Fig. 8. Examples of *Binding* events in the sentences with coordinating conjunctions.

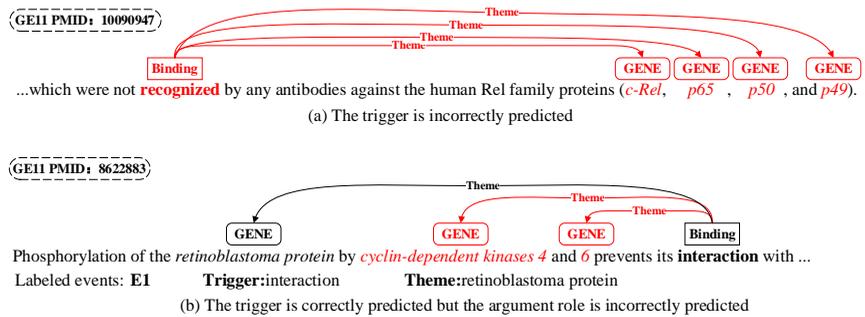

Fig. 9. Examples of cascading errors for *Binding* events.

In summary, the automatic approach takes into account the semantic information of the sentences and thus can exclude the false positives generated by the rule-based approach, leading to the improvement of the precision of *Binding* event extraction.

### 5.2 Recall comparison for Binding event extraction

1) *Recall on GE13 is higher than that on GE11 in the rule-based approach*: As can be seen in Table 5, the recall on the GE13 development set is significantly higher than that on the GE11 development set for *Binding* events under the rule-based approach. This is because the percentage of *Binding* events that are well formed (i.e., a trigger has single argument or two arguments combined to form a *Binding* event) in the rule-based approach is significantly higher in the GE13 development set (about 70%) than in the GE11 development set (about 52%). For example, Fig. 10 shows the common pattern of "binding of A to B" or "A binding to B" in the GE13 development set, which indicates that the binding of *p65*, one of the NFκB protein family, to the repressor protein *IkappaB-alpha*, induces a *Binding* event.

2) *Recall on GE11 is higher than that on GE13 in automatic approach*: As can be seen from Table 5, compared with the rule-based approach, the GE11 recall increases significantly while the GE13 recall decreases, resulting in the former being higher than the latter. The significant increase in the GE11 recall is due to the recall of events that cannot be constructed by the rule-based approach. These are mainly events with only one argument, as shown in Fig. 11(a), where the trigger "**binding**" constitutes a *Binding* event with each argument respectively. On the other hand, the GE13 recall decreases because the pairwise events that can be constructed by the rule-based approach cannot be recognized by the model. As shown in Fig. 11(b), the *Binding* events of the trigger "**binding**" with two arguments are not recognized. The main reason is that the distributions of *Binding* triggers in the GE13 training and development sets are inconsistent, specifically, the domain trigger word of Binding events in the training set is mainly "**recruitment**" instead of "**binding**" in the development set.

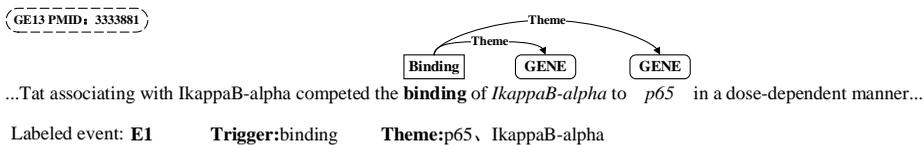

Fig. 10. An example of the pattern of "binding of A to B".

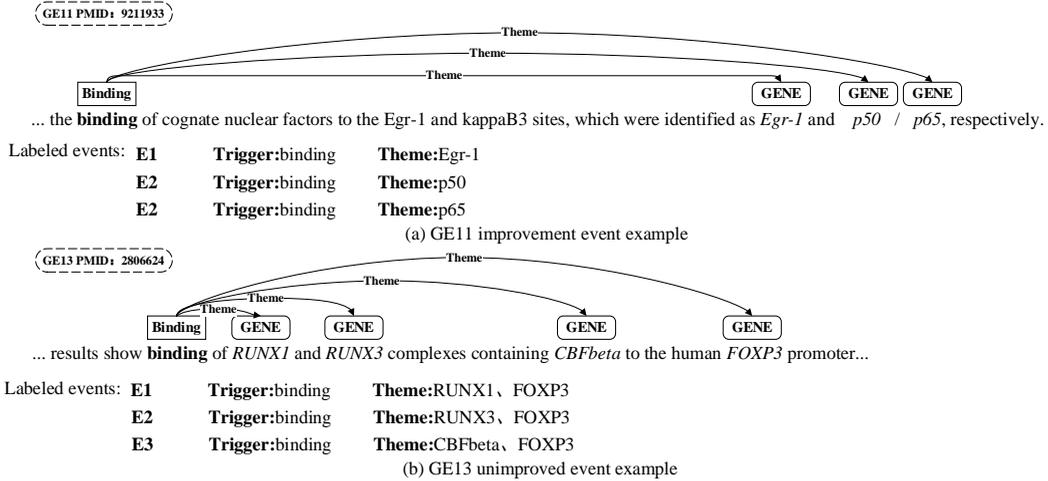

Fig. 11. Examples of *Binding* events in GE11 or GE13.

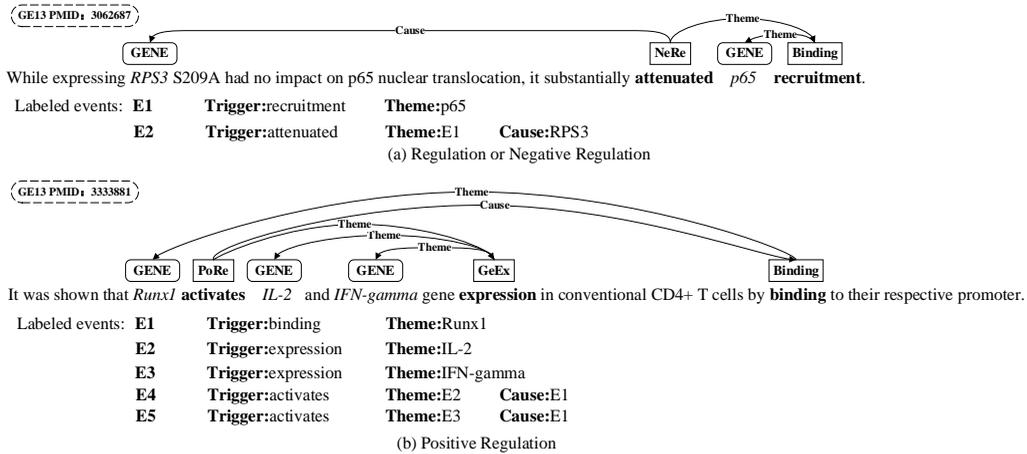

Fig. 12. Examples of regulation events with a *Binding* event as their arguments.

## 5.3 Precision increase of regulation event extraction

As seen in Table 5, the precision scores of all three kinds of regulation event extraction are improved, most significantly for *PoRe*, leading to further improvement of the overall performance of event extraction. Among the three regulation events containing Binding events as arguments, a certain *Binding* event is usually involved in only one *Regu/NeRe* event, as shown in Fig. 12(a), where a *Binding* event is the *Theme* argument of an *NeRe* event. However, a certain *Binding* event is often involved in multiple *PoRe* events especially in the GE13 corpus, as shown in Fig. 12(b), where a *Binding* event is participated in two *PoRe* events. In the rule-based event construction, a false positive of a *Binding* event leads to one false positive of *Regu/NeRe* event, but multiple false positives of *PoRe* event. Statistically, the percentages of false positives of *Regu/PoRe/NeRe* events caused by a wrong *Binding* event as an argument are 0%/46.4%/11% (GE11 development set) and 9.1%/95%/0% (GE13 development set), respectively. In the automatic approach, the false positives of *Binding* events significantly decreased, leading to significant improvements in the precision scores of regulation events, especially the *PoRe* events.

## 6 CONCLUSION AND FUTURE WORK

We perform the biomedical event extraction task in a pipelined manner, with trigger identification, argument role recognition and event construction as sub-tasks, all of which use the BioBERT model as the base encoder. To improve the performance of event construction, we utilize an n-ary relation extraction method to construct *Binding* events, i.e., we classify a candidate event consisting of a combination of triggers and arguments into valid/invalid ones. Experimentation shows that, our method significantly improves the performance of *Binding* events compared with the rule-based approach, and also indirectly improves the performance of regulation events, which finally promotes the event extraction performance under the pipelined approach comparable to or even better than joint models. We also tried to use the n-ary relation extraction method to construct regulation events (*Regu*, *PoRe*, *NeRe*), but since the majority of regulation events with both *Theme* and *Cause* arguments are paired between the arguments of different roles, such events can be satisfied by rule-based construction method. Statistically, such events accounted for 98.5% (GE11 development set) and 100% (GE13 development set) of the regulation events with both *Theme* and *Cause* arguments, respectively, so the final results obtained were not much different compared to those using the rule-based approach.

Our analysis shows that the cascading errors of events due to incorrect trigger prediction account for a larger proportion, so the performance of trigger identification is particularly important in the pipelined approach. In order to exploit the domain-specific knowledge in biomedical event extraction, the use of a machine reading comprehension framework combined with external knowledge for trigger identification may be considered for future work.

## ACKNOWLEDGMENT

This work is supported by the National Natural Science Foundation of China [61976147] and the research grants of The Hong Kong Polytechnic University (#1-W182, #GYW4H) and A Project Funded by the Priority Academic Program Development of Jiangsu Higher Education Institutions (PAPD).